\documentclass[a4paper,fleqn]{cas-dc}

\usepackage{calc,amsmath,amssymb,amsfonts}
\usepackage[LGR,T1]{fontenc}
\usepackage[greek,english]{babel}
\usepackage{xcolor}
\usepackage{enumitem}
\usepackage{hyperref}
\usepackage[authoryear,round]{natbib}

\usepackage{graphicx}
\usepackage{listings}
\usepackage{float}
\usepackage{booktabs} 
\usepackage[most]{tcolorbox} 

\hypersetup{colorlinks=true,allcolors=blue}

\newcommand{\highlightblue}[1]{\textcolor{blue}{#1}}

\lstdefinelanguage{json}{
    morestring=[b]",
    stringstyle=\color{blue},
    literate=
     *{:}{{{\color{red}{:}}}}{1}
      {,}{{{\color{red}{,}}}}{1}
      {\{}{{{\color{black}{\{}}}}{1}
      {\}}{{{\color{black}{\}}}}}{1}
      {[}{{{\color{black}{[}}}}{1}
      {]}{{{\color{black}{]}}}}{1},
}

\newtcolorbox{promptbox}[1][]{
  breakable,
  colback=white,
  colframe=black,
  arc=3mm,
  boxrule=0.5pt,
  fonttitle=\bfseries,
  fontupper=\small, 
  title=#1,
  left=4mm,
  right=4mm,
  top=2mm,
  bottom=2mm
}

\newtcolorbox{appendixpromptstyle}{
  width=\linewidth,
  colback=white,
  colframe=black,
  arc=3mm,
  boxrule=0.5pt,
  breakable,
  left=2mm,
  right=2mm,
  top=3mm,
  bottom=3mm,
  parbox=false,
  before skip=\medskipamount,
  after skip=\medskipamount,
  fontupper=\footnotesize\raggedright 
}

\ExplSyntaxOn
\cs_set:Npn \__first_footerline:
  { \group_begin: \small \sffamily \__short_authors: :~ { \rmfamily \itshape Accepted~manuscript } \group_end: }
\ExplSyntaxOff

\begin{document}

\shorttitle{Machine-Readable Operator Manuals over the ISO 11783 Network for Offline, Multilingual LLM Assistance}
\shortauthors{T. Oksanen and J. Gun}
\title [mode = title]{Machine-Readable Operator Manuals over the ISO 11783 Network for Offline, Multilingual LLM Assistance}
\author[1,2]{Timo Oksanen}\cormark[1]
\author[1]{Julius Gun}
\affiliation[1]{organization={Chair of Agrimechatronics, Technical University of Munich}, country={Germany}}
\affiliation[2]{organization={Munich Institute of Robotics and Machine Intelligence (MIRMI)}, country={Germany}}
\cortext[1]{Corresponding author: timo.oksanen@tum.de}
\nonumnote{This is the accepted manuscript of an article published in \textit{Computers and Electronics in Agriculture}, 112595 (2026). Please cite the published version: \url{https://doi.org/10.1016/j.compag.2026.112595}. \textcopyright~2026. This manuscript version is made available under the CC BY-NC-ND 4.0 license \url{https://creativecommons.org/licenses/by-nc-nd/4.0/}.}

\begin{abstract}
If agricultural machinery is to support intelligent, offline assistance, then the system must access technical documentation dynamically. In this paper, the Semantic Synchronization of operator manuals over ISO 11783-based CAN buses is analyzed, defined as the process that brings the local knowledge base of a tractor-side assistant into agreement with the operator manual of the connected implement. Data transfer performance was benchmarked using Markdown, JSON, and XML; various open-weight Large Language Models (LLMs) were also evaluated in cross-lingual query tasks. The automatic evaluation of the answers was validated against four human raters. A complete manual was transferred over a physical ISO 11783 network in 31 to 89 s, depending on format and transport settings, which is comparable to the object pool upload that operators already wait for at the first coupling of an implement. Markdown produced the smallest files, while the question-answering performance was comparable across the three formats for most models. With retrieval-augmented generation, locally run models answered most questions on the English manual correctly, with no advantage of the largest model over 7B and 8B models, but performed worse on the five other language editions. Keyword retrieval failed across languages, whereas semantic retrieval provided the cross-lingual capability. The exchange requires only a new parameter group that uses the existing request and transport mechanisms of ISO 11783, which makes it a candidate for standardization. The study is a proof of concept with inference on PC-grade hardware; deployment on embedded agricultural hardware remains future work.
\end{abstract}


\maketitle

\section{Introduction}

Integrating Large Language Models (LLMs) into agricultural machinery requires a robust solution for offline data access. Printed operator manuals cannot provide in-situ support in the tractor cabin because they are difficult to navigate during operations. Although researchers have investigated these LLMs for office and mobile support, applications now extend to embedded systems. To integrate these systems into agriculture, however, we must navigate the specific constraints of the standardized vehicle network.

Agricultural machinery in Europe features multi-brand systems where farmers connect implements to the tractor in a plug-and-play manner using ISO 11783 \citep{ISO11783-1}. Farmers supplement the connected system with retrofit electronics, such as GNSS receivers and user terminals. ISO 11783 (ISOBUS) is the standard that interconnects the tractor, implement, user interface, farm management system, and precision farming task controller. This interconnected environment provides the necessary infrastructure for advanced data exchange.

In this paper, we explain how a local LLM-based assistant in an ISO 11783 system accesses textual data from other subsystems in an open network. Specifically, we examine machine-readable digital operator manuals stored in an implement's Electronic Control Unit (ECU) and accessed by a tractor terminal in the tractor cabin. Access to these data is crucial in rural areas where internet connections are unreliable. Storing the manual in the implement ECU keeps it with the machine throughout its lifecycle. This study is a proof of concept: we evaluate the transfer on a physical ISO 11783 network and the question answering with locally run open-weight LLMs on PC-grade hardware, as a step toward deployment on resource-constrained ECUs. Unlike autonomous driving, where AI actuates the vehicle, the proposed system operates strictly as an information retrieval layer (Level 0 autonomy), which places the human operator as the sole decision maker.

The ISO 11783 series standardizes the user interface (Virtual Terminal, \citealt{ISO11783-6}), task data exchange (Task Controller, \citealt{ISO11783-10}) and diagnostics limited to version information, capabilities and I/O errors \citep{ISO11783-12}, but it provides no mechanism by which an implement can supply its technical documentation to a local assistant on the tractor side, and existing retrieval-augmented systems rely on centrally maintained document collections \citep{IEEE11204203, koopman2022agaskagenthelpanswer}. A standardized exchange is needed, which would allow tractor manufacturers to build their own assistants and implement manufacturers to provide their own serialization of the manual. Following the layered ISO/OSI architecture of ISO 11783, such an exchange should be added at the application layer as a new parameter group that uses the existing network management, request and extended transport mechanisms unchanged, rather than as a proprietary extension. We define \textit{Semantic Synchronization} as the process that brings the local knowledge base of the tractor-side assistant into agreement with the manual of the connected implement, as in data synchronization: its input is the versioned manual stored in the implement ECU, its output a complete local copy together with the retrieval index built from it; it succeeds when the manual is received completely, its version matches the one announced by the implement and the index is built from it. In this study, the transfer and the indexing were implemented and evaluated, whereas the version check, the first-coupling protocol, the background image transfer and the source-verification interface are proposed design elements. Building on our preliminary study \citep{oldpaper}, we reuse its manual, question-answer dataset, prompts and retrieval configuration; the transfer over ISO 11783, the format comparison, the six language editions and the current models are new. This ManualSync project aims for standardized exchange of machine-readable operator manuals over ISO 11783 for offline, multilingual LLM assistance. We address three \textbf{research questions}:
\begin{enumerate}[label=\arabic*.]

    \item Can a complete operator manual be transferred over ISO 11783 with the existing request and extended transport mechanisms within a practical time at the first coupling of an implement?
    \item How do Markdown, JSON and XML compare in size, transfer time, token count and question-answering performance?
    \item Can locally run open-weight LLMs answer operator questions from the transferred manual, and also across languages? What does each retrieval component contribute?
\end{enumerate}

\section{Motivation}

\begin{figure*}[pos=tp]
    \centering
    \includegraphics[width=\textwidth]{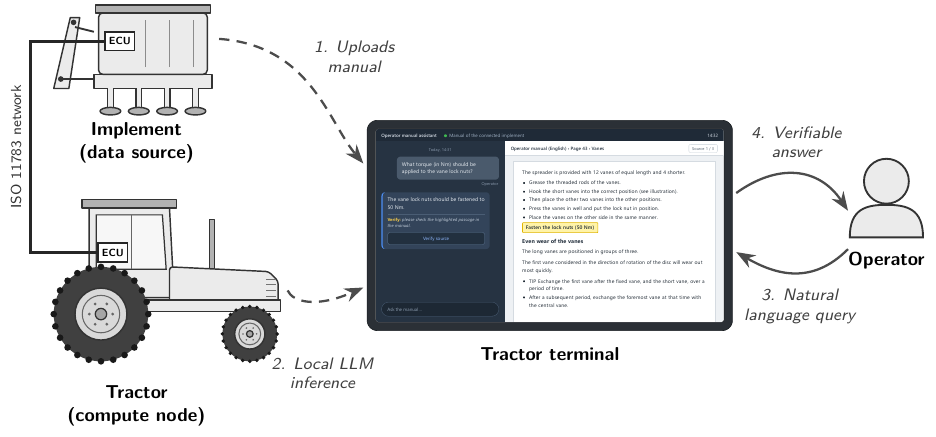}
    \caption{Conceptual architecture for offline Semantic Synchronization. }
    \label{fig:1}
\end{figure*}

We propose a self-describing architecture to bridge the gap stated above. In our design, the implement stores its own documentation in a structured format and transmits these data to the local LLM of the tractor via the ISOBUS network (Figure \ref{fig:1}). This data flow allows the operator to maintain conversational access to safety information regardless of cellular coverage.

\section{State of the Art}

\subsection{Applications of LLMs}
Researchers have recently integrated LLMs into automotive and heavy-equipment domains to enhance user interaction. In autonomous driving, LLMs improve decision-making by combining multimodal sensor inputs \citep{li2024humanlike}, while in the intelligent cockpit, they enable natural-language control of vehicle functions \citep{Jafarnejad2024}. These models predict driver intent accurately and provide a superior user experience compared to conventional menu-driven interfaces \citep{Chen2025Cockpit}. In the specific context of agriculture, \citet{johnson2024enhancing} showed that multi-round prompt engineering significantly improves machinery management. Multimodal assistants such as Agri-LLaVA address agronomic questions on pests and diseases \citep{wang2024agrillavaknowledgeinfusedlargemultimodal}. 

Recent studies also investigate architectures where large language models interact with structured technical information or domain knowledge sources in industrial systems \citep{IEEE11204203, IEEE11087302}. Although these works explore LLM-driven knowledge interfaces, our work distinguishes itself by enabling such functionality directly over the ISO 11783 (ISOBUS) machinery network. The domain has specific requirements for hot plug-and-play functionality, making ISOBUS the fundamental constraint for our system design.

\subsection{ISO 11783 Transfer Mechanism}

To move the data required by these models, it is necessary to look to the existing transport layers of the vehicle network. The ISO 11783 standard series provides the Extended Transport Protocol (ETP) for transmitting extra-large data messages over the CAN bus. The classic Transport Protocol (TP), originally developed jointly for the SAE J1939 and ISO 11783 series, limits message transmission to 1785 bytes. ETP overcomes this constraint by supporting messages up to approximately 117 million bytes through enhanced packet addressing capabilities. ETP uses two distinct Parameter Groups: ETP.CM (Connection Management, PGN=51200) for control messages and ETP.DT (Data Transfer, PGN=50944) for payload transmission \citep{ISO11783-3}.

The protocol manages the connection through a specific handshake sequence. ETP operates through connection-oriented sessions where the originator initiates the session by sending a Request to Send (RTS) message that specifies the total message size and the target PGN. If ready, the receiver responds with a Clear to Send (CTS) message to indicate reception capacity. Although 16 is the industry default, this capacity supports 255 packets per burst. Each data packet contains seven bytes of payload and the sequence number. If the receiver detects packet loss, it requests retransmission using Data Packet Offset commands. An End-of-Message Acknowledgement confirms session completion, whereas abort messages terminate a session abnormally.

This protocol manages the 8-byte limitation of the CAN bus, making it the primary candidate for transferring large documentation files.

\subsection{Cross-lingual Capability in Machinery}
Because agricultural machinery operates across diverse linguistic regions, manufacturers must provide operator manuals in multiple languages. \citet{chua2025crosslingual} evaluated LLMs on cross-lingual tasks and found that although models show promising surface-level abilities, they struggle with deep cross-lingual knowledge transfer. Their work demonstrated that fine-tuning on mixed-language data effectively reduces these performance gaps.

Recent agricultural studies confirm the value of multilingual access. \citet{kaur2025leveraging} generated multilingual synthetic datasets from agriculture-specific documents and achieved improvements in factuality by fine-tuning LLMs for question answering across English, Hindi, and Punjabi. Similarly, \citet{Samuel2025AgroLLMCF} demonstrated that combining LLMs with RAG frameworks achieves high accuracy for agricultural knowledge sharing. These findings suggest that an LLM with cross-lingual capabilities could answer user queries in one language using documentation stored in another, reducing the need to store multiple translations in the resource-constrained memory of an ECU.

\subsection{Retrieval-Augmented Generation (RAG)}
Retrieval-Augmented Generation (RAG) offers a solution to the limited knowledge base of pretrained LLMs. \citet{vizniuk2025comprehensive} identify RAG as a key approach for overcoming the limited knowledge base of pretrained LLMs in agricultural decision-support systems. At the retrieval level, \citet{saha2024quimrag} match user questions against questions generated from the target corpus to reduce information dilution and hallucination in RAG-based question answering.

Several systems already follow this pattern. AgAsk \citep{koopman2022agaskagenthelpanswer} answers farmers' natural-language questions by retrieving relevant passages from scientific documents, effectively implementing an early RAG-style pipeline for agriculture-focused QA. PooRaa-Agri KG \citep{article} uses a knowledge graph and a multilingual query interface to return concise answers in English and Hindi, separating retrieval over a structured domain representation from the answer layer. AgroLLM \citep{Samuel2025AgroLLMCF} explicitly combines a vector database (FAISS) with LLMs in a RAG framework. The authors report substantial accuracy gains over non-RAG baselines across several agricultural subdomains.

At the architectural level, \citet{11080430} review RAG-driven memory designs for conversational LLMs. They highlight how vector databases can integrate long-term domain knowledge with user-specific data. In parallel, recent work shows that modern tractors and implements already expose rich digital traces via CAN bus and GNSS \citep{agriengineering1010010,BETTUCCI2024100511}. These strands suggest that RAG architectures are well suited for linking machinery data to technical documentation.

\subsection{Large Context Window}
Recent advancements in LLMs allow models to process entire documents simultaneously, but this capability has limits. Models like GPT-OSS 20B and Gemma 3 12B now support context windows of up to 128k tokens. \citet{liu2023lostmiddlelanguagemodels}, however, describe the \textit{lost-in-the-middle} effect, where models fail to retrieve information located in the middle of long contexts. It was found that small models perform better if the system uses RAG to supply only a small, relevant context \citep{oldpaper}. This approach avoids the degradation associated with processing massive documents directly.

\section{Materials}

\subsection{Operator Manual Structure}

Under EU Machinery Directive 2006/42/EC \citep{EU2006}, agricultural machinery manufacturers must provide a comprehensive operator manual to meet health and safety requirements. Manufacturers typically achieve compliance through the harmonized standard EN ISO 4254-1 \citep{ISO4254-1}, which sets requirements for the content of the operator manual. These requirements include mounting and dismounting procedures, tractor compatibility specifications, stability information, control descriptions with standardized symbols, ergonomic adjustments, starting and stopping procedures, and emergency exits. Additionally, the manual must cover moving-parts precautions, maintenance procedures, noise and vibration emission values, energy hazards (battery and fuel), lifting points, and power take-off shafts.

These regulations create structural homogeneity across operator manuals from different manufacturers. Because regulations mandate several chapters in the operator manual, the document organization is predictable, even if not unified. This standardization driven by legal compliance creates an interesting foundation for machine-readable documentation systems because information categories may allow automated content extraction, classification, and analysis. We selected the operator manual of an agricultural fertilizer spreader because its structure is shared by manuals of similar machines under the scope of this regulation. Fertilizer spreaders have their own standard for safety and health, ISO 4254-8 \citep{ISO4254-8}. 

\subsection{Selected Operator Manual}
\label{sec:selected_manual}
Our test case is the operator manual for the Kverneland Exacta-TLX Geospread GS3, a mechatronic fertilizer spreader. We chose this manual because our familiarity with the machine aided in creating the handcrafted QA set. This machine also met our requirement for operator manuals publicly available without authentication in numerous European languages.

The manuals were used in their official editions: English \citep{KvernelandManualEN}, French \citep{KvernelandManualFR}, German \citep{KvernelandManualDE}, Italian \citep{KvernelandManualIT}, Dutch \citep{KvernelandManualNL}, and Spanish \citep{KvernelandManualES}. We verified that all six editions refer to the same machine and software version. Each 165-page manual possesses an identical layout across languages; the Markdown version contains approximately 61k tokens in English and 83k to 89k tokens in the other languages (Qwen3-Embedding-8B tokenizer). The number of words varies by language: English (41094), Dutch (37972), French (46047), German (38041), Italian (43201), and Spanish (45751).

\begin{figure}[pos=htbp]
    \centering
    \includegraphics[width=0.6\columnwidth]{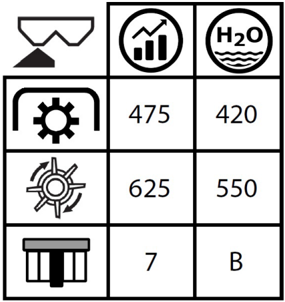}
    \caption{Representative image from the operator manual (page 86) illustrating PTO speeds for boundary spreading.}
    \label{fig:manual_page_86}
\end{figure}

How images affect data exchange was also explored. Because the manual was not designed for machine readability, one representative image was evaluated for benchmarking. This image, shown in Figure \ref{fig:manual_page_86}, is from page 86 of the manual and illustrates a graphical table of PTO speeds for various boundary spreading settings. Most graphics in this manual follow a similar style. 

Because ISO 11783-6 already provides a presentation layer for picture graphics, the same compression method was used. The sample image is natively 578 pixels wide and 610 pixels high, resulting in 352580 bytes in the 256-indexed palette used in ISO 11783-6. Two compression methods can be enabled for this image: reduction to a 16-indexed color palette and run-length encoding (ISO 11783-6, section B.12). Run-length encoding compresses repeating sequences of the same value with two-byte pairs. We selected this method because the ISO 11783 Virtual Terminal server software implements it, and it is industry-proven and lightweight. Together, the 16-indexed color palette and run-length encoding provide a compression baseline for the following benchmark.

\subsection{QA Test Dataset}
The QA dataset from our previous study \citep{oldpaper}, which comprises 108 questions curated from domain expertise, was used. The complete dataset is available in the associated code repository. The dataset is balanced with 54 answerable questions targeting specific operational and safety information (e.g., torque specifications, maintenance intervals) and 54 unanswerable questions designed to test model hallucination tendencies.

Unanswerable questions are semantically plausible but cannot be answered from the manual content (e.g., "What is the minimum baud rate for RS 232 connection?"). This design enables us to evaluate both retrieval accuracy and the model's ability to appropriately decline answering when information is absent—a critical capability for safety-relevant applications. Following established benchmarks like XTREME \citep{hu2020xtrememassivelymultilingualmultitask}, all questions were posed in English regardless of the document language to isolate cross-lingual retrieval capability. Example questions from both categories are provided in Appendix~\ref{sec:appendix_qa_dataset}.

\section{Methods}
We converted the publicly available PDF manuals into Markdown, JSON, and XML using the Docling library \citep{Docling} and developed a wrapper to enable page-wise conversion. The wrapper converts each PDF once with the default Docling PDF pipeline (Docling version 2.26) and serializes the recognized items page by page into the three formats: section headers, text and list items as text; tables as Markdown tables, JSON objects or fixed-width text in XML; and pictures, including safety pictograms, as placeholders with their caption where one was recognized. Page headers and footers were not included. The converted files were used without manual editing and were stored as UTF-8 text with Windows line endings (CRLF), as transferred.

\subsection{Semantic Synchronization}
\label{sec:semsync_methods}
To evaluate the transfer of manual data over the vehicle network, we established a test setup using the ISO 11783 Extended Transport Protocol (ETP). We used a temporary proprietary PG number (65280) to prove the concept, together with a standard request message (PGN 59904) and the existing ETP mechanism. The priority of messages was set to 7 (lowest), which is the default for ETP. 

We created the test setup with a physical CAN bus and two connected computers. One software application acted as the operator manual server and the other as the operator manual client. During the test, the physical CAN bus backbone also ran other control functions, namely an AEF-certified commercial Virtual Terminal, task controller, and file server, to provide a realistic plug-and-play system. Without the manual transfer, the average background bus load from these control functions was below 2\%. At 250 kbit/s \citep{ISO11783-2}, the bus carries approximately 1900 frames per second (131 bits per 8-byte extended frame, excluding bit stuffing). This background corresponds to the condition at the first connection of an implement, when the system is stationary and the operator waits for the object pool upload \citep{ISO11783-6}. We aimed to ensure that the normal operation of the plug-and-play network stays fully functional despite the additional burst of this data exchange. 

In the proposed design, the system does not perform this transfer during every Key-On cycle. Instead, it uses a first-coupling protocol: the manual is transferred only when a new implement is attached for the very first time and is stored in the tractor's non-volatile memory. To handle implement firmware updates that might alter the manual, the system employs a version check. The ECU checks the version of the manual on every startup; a transfer only occurs if it differs from the cached version. Subsequent connections require only this check. This follows the existing pattern of ISO 11783-6, in which the Virtual Terminal can store the object pool of an implement under a version label, so that the pool is uploaded only at the first connection or after a change \citep{ISO11783-6}; device descriptions of the task controller, which are transferred at the same time, take precedence over the manual.

For Semantic Synchronization, the system serialized each operator manual into a byte array. This serialization preserves encoding but raises the question of the appropriate structural format: JSON, XML, or Markdown. Because we have six languages and three structural formats, we tested 18 combinations. We measured the time required to transfer the complete manual in different formats to determine the most bandwidth-efficient solution. The system also measured the bus load from the server side by accessing the hardware-level diagnostics counters, which account for bit stuffing.

\subsection{RAG Configuration}
We used a Hybrid RAG retrieval strategy. We detail the ablation study of the retrieval configuration in Appendix \ref{sec:appendix_ablation}. This strategy combines BM25 keyword search with semantic retrieval using the \texttt{Qwen3-Embedding-8B} model. To merge the results, we used Reciprocal Rank Fusion (RRF) \citep{RRF}, which produces a robust final ranking by prioritizing chunks that rank highly in both methods. We split documents into 200-token segments with a 100-token overlap and retrieved the top three chunks per query. Chunks were created with the tokenizer of the embedding model, which ran on the CPU in 32-bit floating point. BM25 (Okapi variant) used lowercase whitespace tokens; the semantic search used ChromaDB with L2 distance on normalized embeddings, which ranks identically to cosine similarity. Each retriever returned ten candidates, which were fused with RRF ($k = 60$); the top three fused chunks were passed to the model. This configuration was chosen on the basis of our preliminary study \citep{oldpaper} and is re-evaluated in Appendix~\ref{sec:appendix_ablation}.

\subsection{Models}
\label{sec:scoring}

We evaluated a range of open-weight LLMs deployed locally via Ollama (Table~\ref{tab:model_overview}). All models were evaluated with a temperature of 0, without a fixed seed, with an output limit of 1024 tokens and with the default reasoning setting of each model. Confidence intervals were estimated by question-level bootstrap resampling (2000 resamples, stratified by answerable and unanswerable questions; questions resampled as clusters when editions were pooled), and paired comparisons used a paired bootstrap with Holm correction within each family of comparisons (formats; context conditions) (Appendix~\ref{sec:appendix_results}).

Model sizes range from 1.5B to 120B parameters, which enables analysis of the trade-off between model capacity and performance. All models were executed with 4-bit quantization (Q4\_K\_M; MXFP4 for GPT-OSS), which reduces the model weights of the 8B-parameter models to about 5 GB. 

For automatic evaluation, we used an LLM-as-a-judge framework \citep{zheng2023judging} with GPT-OSS 20B as the evaluator across all experiments, ensuring consistency. The judge compared model outputs against ground-truth answers using the prompts in Appendix~\ref{sec:appendix_prompts}. If the judge returned no valid verdict (147 of 15552 items, 0.9\%; mostly because its reasoning reached the output limit), the call was repeated with identical settings (123 items resolved) and, for the remaining 24 items, with the reasoning effort set to low. A verdict of yes counted as a true positive for answerable and as a true negative for unanswerable questions, and a verdict of no as a false negative or a false positive, respectively; answers written in the language of the manual edition instead of English counted as incorrect. The language of each answer was determined with an automatic language detector (lingua); answers mixing English with the language of the manual (226) were scored by the verdict of the judge. The verdicts of the judge were validated against four independent human raters, with whose majority the judge agreed for 93.5\% of the items (Cohen's $\kappa = 0.87$); its errors were mainly rejections of acceptable answers, so the reported F1 scores are conservative (Appendix~\ref{sec:appendix_judge_validation}).

\begin{table*}[pos=tp]
    \centering
    \caption{Overview of LLMs evaluated in this study.}
    \footnotesize
    \label{tab:model_overview}
    \begin{tabular}{lrlrl}
        \toprule
        \textbf{Model} & \textbf{Size} & \textbf{Ollama tag} & \textbf{Parameters} & \textbf{Quantization} \\
        \midrule
        GPT-OSS & 120B & gpt-oss:120b & 116.8B & MXFP4 \\
        GPT-OSS (judge) & 20B & gpt-oss:20b & 20.9B & MXFP4 \\
        Gemma 3 & 12B & gemma3:12b & 12.2B & Q4\_K\_M \\
        Qwen 3 & 8B & qwen3:8b & 8.2B & Q4\_K\_M \\
        DeepSeek-R1 & 8B & deepseek-r1:8b & 8.2B & Q4\_K\_M \\
        Qwen 2.5 & 7B & qwen2.5:latest & 7.6B & Q4\_K\_M \\
        Llama 3.1 & 8B & llama3.1:8b & 8.0B & Q4\_K\_M \\
        Llama 3.2 & 3B & llama3.2:3b & 3.2B & Q4\_K\_M \\
        DeepSeek-R1 & 1.5B & deepseek-r1:1.5b & 1.8B & Q4\_K\_M \\
        \bottomrule
    \end{tabular}
\end{table*}
\subsection{Formatting Analysis}
\label{sec:formatting_analysis}

To evaluate the efficiency of data transfer and ingestion, we selected three formats representing distinct data interchange paradigms. We explicitly excluded unstructured plain text because it fails to preserve the document hierarchy (headers, lists, warnings) that accurate RAG retrieval requires. We also excluded HTML because of the security risks associated with rendering web content in safety-critical vehicle displays. Furthermore, we excluded compressed office formats (e.g., DOCX) because the decompression process consumes excessive CPU cycles during the critical machine startup phase.

We excluded binary serialization formats (e.g., Protobuf) although they save bandwidth. We argue that binary formats introduce a schema coupling dependency: they require both the transmitter (implement) and receiver (tractor) to possess identical schema definitions (e.g., .proto files) to decode the payload. This dependency violates the plug-and-play core principle of ISOBUS; a tractor would require a firmware update to read documentation from a new implement. In contrast, the text-based formats below are self-describing and require no prior schema exchange.

\begin{itemize}
    \item \textbf{XML (The Legacy Standard)}: XML represents the incumbent standard for agricultural and automotive data (e.g., ISO 11783-10 TaskData, ODX). The XML files use a flat structure of page, text, table and image elements without item types.
    \item \textbf{JSON (The Web/API Standard)}: JSON represents the modern standard for telematics and cloud connectivity. The JSON files list the items of each page as objects with their item type (e.g., section header, text, list item) and are indented by two spaces.
    \item \textbf{Markdown (The LLM-Native Standard)}: We selected Markdown as the representative stream-based format. It offers a content-first approach, using minimal syntax characters (\#, *, -) to denote structure rather than explicit opening/closing tags or key-value pairs. Markdown provides a good balance: it achieves native token efficiency without sacrificing immediate human verification.
\end{itemize}

\newsavebox{\boxMD}
\begin{lrbox}{\boxMD}
\begin{minipage}[t]{0.3\textwidth}
\begin{lstlisting}[basicstyle=\tiny\ttfamily, breaklines=true, frame=none, aboveskip=0pt, belowskip=0pt]
## Machine identification 
 
In order to provide you with good service various details about your machine may sometimes be required. Please complete these details here in order to have them at hand when they are requested. 
\end{lstlisting}
\end{minipage}
\end{lrbox}

\newsavebox{\boxJSON}
\begin{lrbox}{\boxJSON}
\begin{minipage}[t]{0.3\textwidth}
\begin{lstlisting}[language=json, basicstyle=\tiny\ttfamily, breaklines=true, frame=none, aboveskip=0pt, belowskip=0pt]
{  "page_number": 2, 
  "items": [    { 
      "type": "SectionHeaderItem", 
      "text": "Machine identification"  },   { 
      "type": "TextItem", 
      "text": "In order to provide you with good service various details about your machine may sometimes be required. Please complete these details here in order to have them at hand when they are requested." },    
\end{lstlisting}
\end{minipage}
\end{lrbox}

\newsavebox{\boxXML}
\begin{lrbox}{\boxXML}
\begin{minipage}[t]{0.3\textwidth}
\begin{lstlisting}[language=xml, basicstyle=\tiny\ttfamily, breaklines=true, frame=none, aboveskip=0pt, belowskip=0pt]
<page number="2"> 
    <text>Machine identification</text> 
    <text>In order to provide you with good service various details about your machine may sometimes be required. Please complete these details here in order to have them at hand when they are requested.</text> 
\end{lstlisting}
\end{minipage}
\end{lrbox}

\begin{table*}[pos=tp]
    \centering
    \caption{Example of different formatting of text on page 2.}
    \label{tab:formatting_examples}
    \begin{tabular}{p{0.30\textwidth}p{0.30\textwidth}p{0.30\textwidth}}
        \toprule
        \textbf{Markdown} & \textbf{JSON} & \textbf{XML} \\
        \midrule
        \usebox{\boxMD} & \usebox{\boxJSON} & \usebox{\boxXML} \\
        \bottomrule
    \end{tabular}
\end{table*}
Table \ref{tab:formatting_examples} presents examples of Markdown, JSON, and XML formats. We compared these formats for the constrained environment of agricultural machinery.

\subsection{Cross-lingual Tests}
We assessed the models' ability to retrieve information across languages by posing English questions against manuals in German, French, Dutch, Italian, and Spanish. This test simulates a scenario where a user queries the system in their native language (English in this test case) while the machine's internal documentation is stored in the manufacturer's original language or another available translation. We deliberately selected this cross-lingual retrieval task (English Query $\rightarrow$ Source Language Document) to isolate cross-lingual question answering, following established benchmarks like XTREME \citep{hu2020xtrememassivelymultilingualmultitask}.

\section{Results}

\subsection{Semantic Synchronization}
\label{sec:semsync_results}

The prototype implementation on the physical CAN bus of ISO 11783 ETP transfer revealed significant differences in transmission latency. We tested all combinations of six languages (Dutch, French, English, German, Italian, and Spanish) and three text structural encodings (Markdown, XML and JSON), plus a few more variants of the ETP parameter CTS. See Table \ref{tab:etp_results} in Appendix \ref{sec:appendix_sequence_protocol} for detailed results about transfer times and used data packets on the CAN bus. 

\begin{figure}[pos=htbp]
    \centering
    \includegraphics[width=\columnwidth]{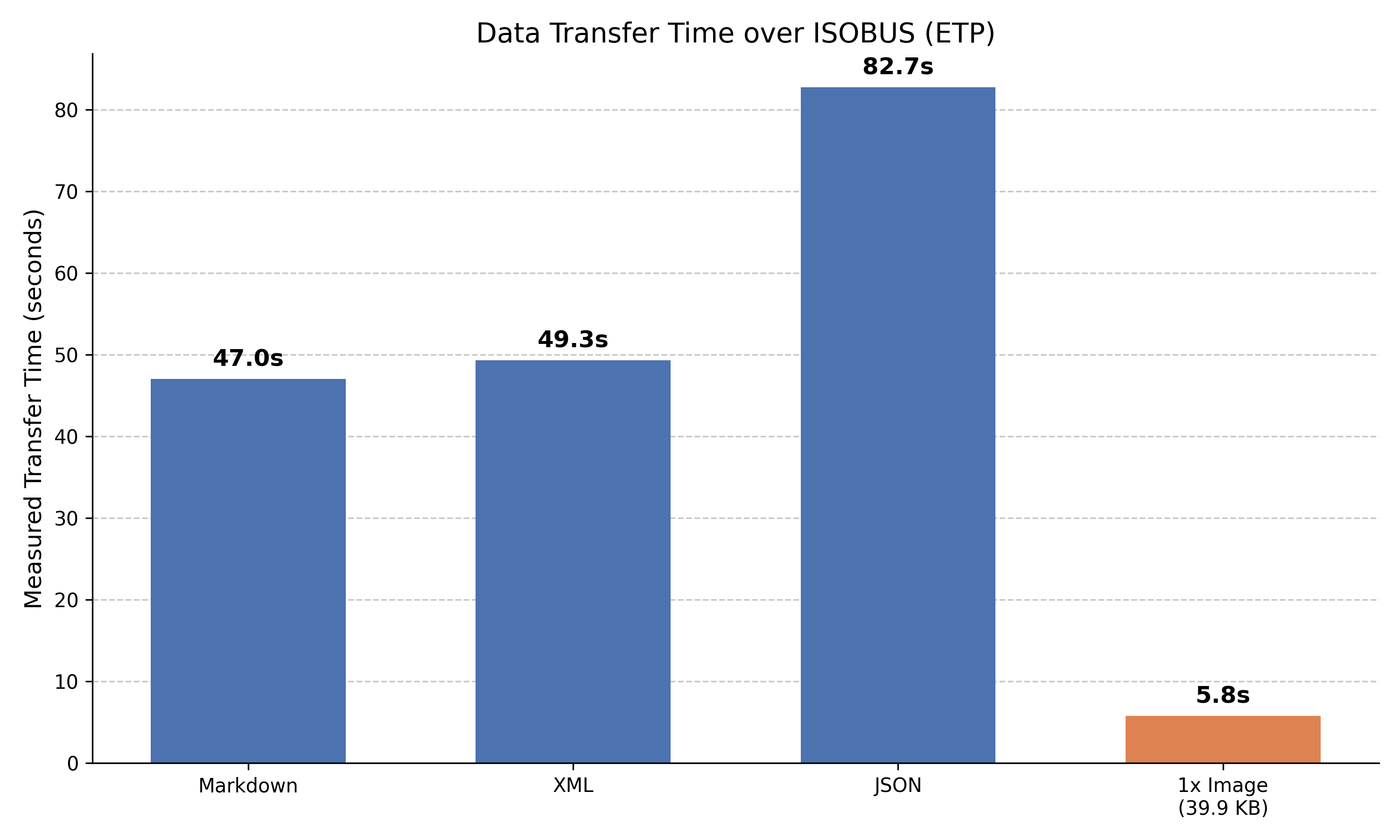}
    \caption{Comparison of measured data transfer time over ISOBUS for different manual formats for the English manual (CTS 32).}
    \label{fig:transfer_time}
\end{figure}

\noindent Summary of experimental results: 

\begin{itemize}
    \item \textbf{Markdown:} Markdown produced the smallest files, leading to a transfer time of 47 s (at CTS 32; Figure~\ref{fig:transfer_time}).
    \item \textbf{XML:} XML introduced approximately 5\% overhead compared to Markdown.
    \item \textbf{JSON:} The indented JSON files caused the highest overhead (75\% increase); without indentation, the English JSON file would be 351166 bytes instead of 556063.
\end{itemize}

The transmission of visual data of graphics typical for this operator manual was also tested (Figure \ref{fig:manual_page_86}). For the data exchange experiment, the run-length encoded bitmap (39855 bytes), which is defined in ISO 11783-6, was used. Transferring these graphics with this encoding took 5.8 s.

The transfer times were measured at a background bus load below 2\%, which corresponds to the first coupling of an implement (Section~\ref{sec:semsync_methods}).

\subsection{Formatting Efficiency}
\begin{figure}[pos=htbp]
    \centering
    \includegraphics[width=\columnwidth]{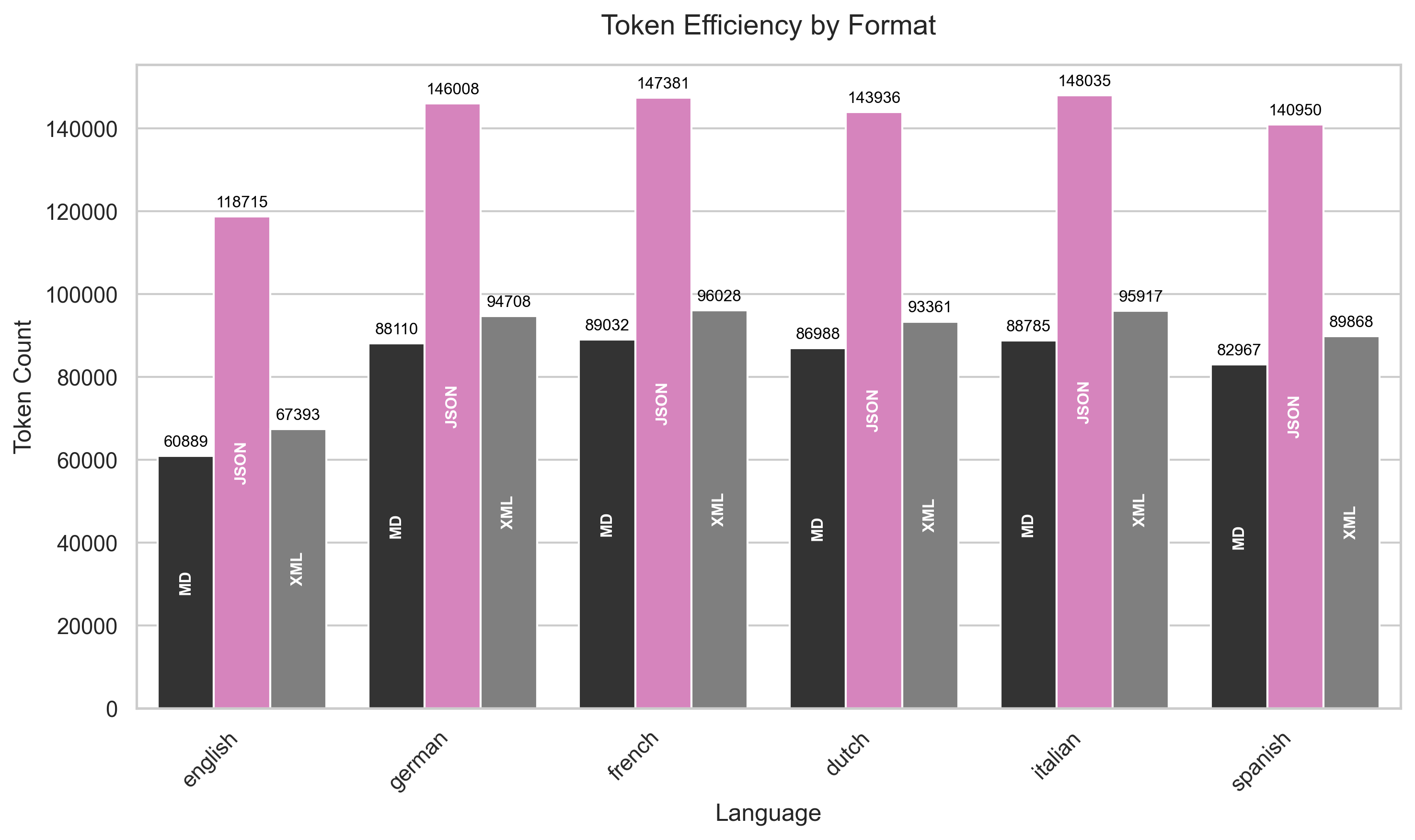}
    \caption{Token efficiency comparison: XML vs JSON vs Markdown across different languages.}
    \label{fig:token_counts}
\end{figure}

Format choice affects transmission time and context window usage. Markdown was the most token-efficient format for ingestion. The English version required approximately 30\% fewer tokens than the other European languages (Figure \ref{fig:token_counts}). This difference results from the tokenizer, which splits non-English text into more tokens; the word counts of the editions are similar (Section~\ref{sec:selected_manual}). Formatting choice significantly affects token count. The JSON format adds substantial overhead, increasing token usage by approximately 95\% compared to Markdown in the English dataset. We note that token efficiency varies by model architecture and tokenizer implementation; models optimized for non-Latin scripts may exhibit different compression ratios.

\begin{figure*}[pos=tp]
    \centering
    \includegraphics[width=\textwidth]{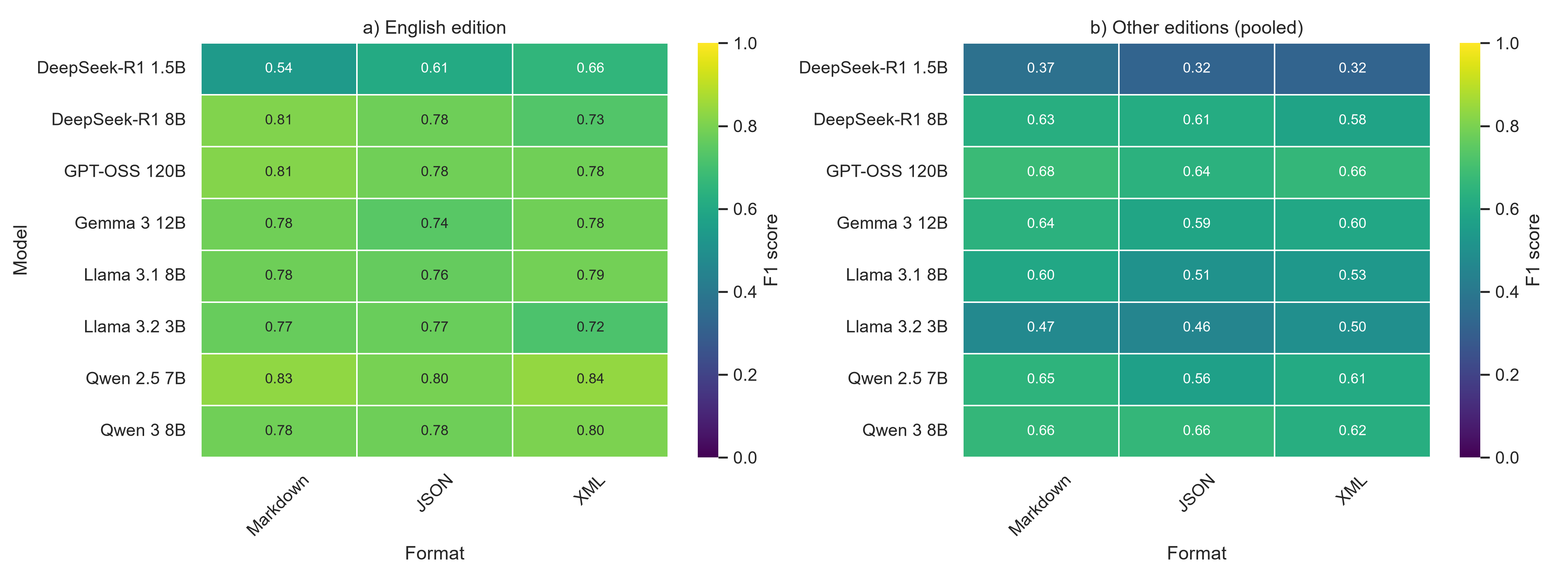}
    \caption{F1 score by model and format: a) English edition, b) five other editions pooled.}
    \label{fig:format_comparison}
\end{figure*}

\subsection{Performance in LLM Tasks}
\label{sec:llm_performance}
We report model performance as F1 score; accuracy, precision, recall and specificity can be derived from the outcome categories in Table~\ref{tab:error_categories}. In our evaluation, a \textbf{positive case} corresponds to an answerable question, and a \textbf{negative case} corresponds to an unanswerable question. This distinction allows us to assess not only correctness but also the ability of the model to avoid hallucination (measured via specificity).

\begin{figure*}[pos=tp]
    \centering
    \includegraphics[width=\textwidth]{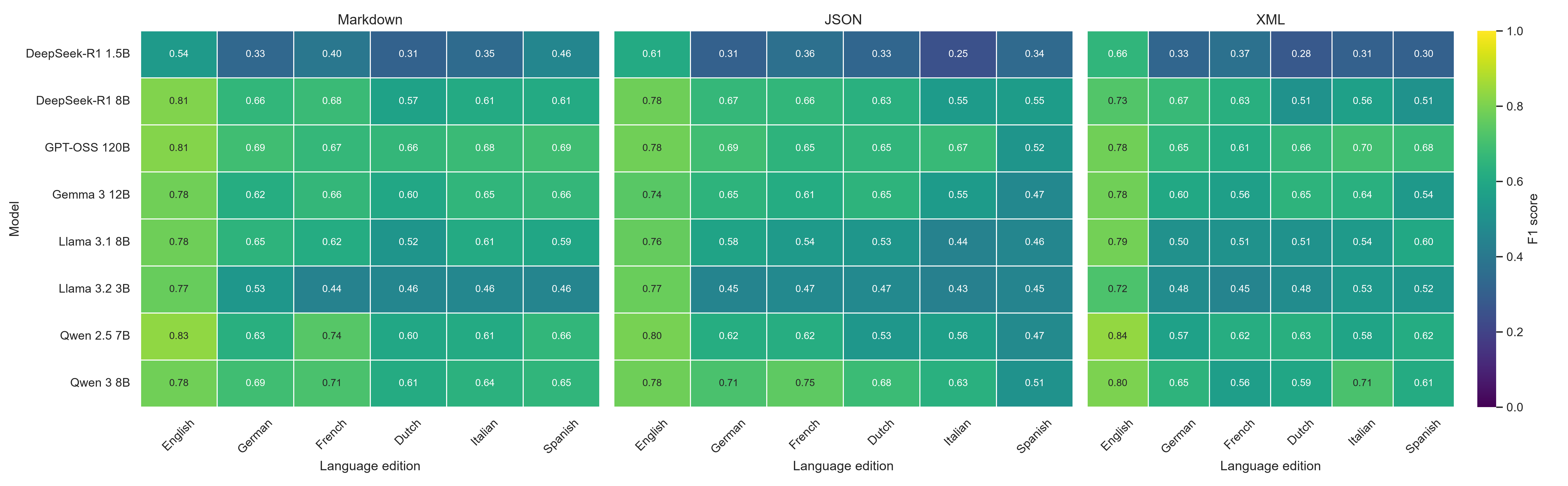}
    \caption{F1 score by model and language edition for the three formats.}
    \label{fig:global_f1}
\end{figure*}

As shown in the global overview (Figure~\ref{fig:global_f1}), performance varied by model size and architecture.

Figure~\ref{fig:format_comparison} shows the F1 score of each model for the three formats, a) for the English edition and b) for the five other editions pooled. For the English edition, the formats differed by at most 0.05 for six of the eight models and by up to 0.12 for the two DeepSeek-R1 models, and none of these differences was significant. For the other editions, Markdown exceeded JSON for Llama 3.1 8B (+0.09) and Qwen 2.5 7B (+0.08), whereas all other format differences were not significant; confidence intervals are given in Appendix~\ref{sec:appendix_results}.

For the English edition with Markdown, F1 ranged from 0.54 (DeepSeek-R1 1.5B) to 0.83 (Qwen 2.5 7B); the seven models from 3B parameters upward reached 0.77 to 0.83 with overlapping confidence intervals, and GPT-OSS 120B (0.81) did not exceed the 7B and 8B models (Appendix~\ref{sec:appendix_results}).

\subsection{Feasibility for Cross-lingual Tasks}
\label{sec:crosslingual_results}
\begin{figure*}[pos=tp]
    \centering
    \includegraphics[width=\textwidth]{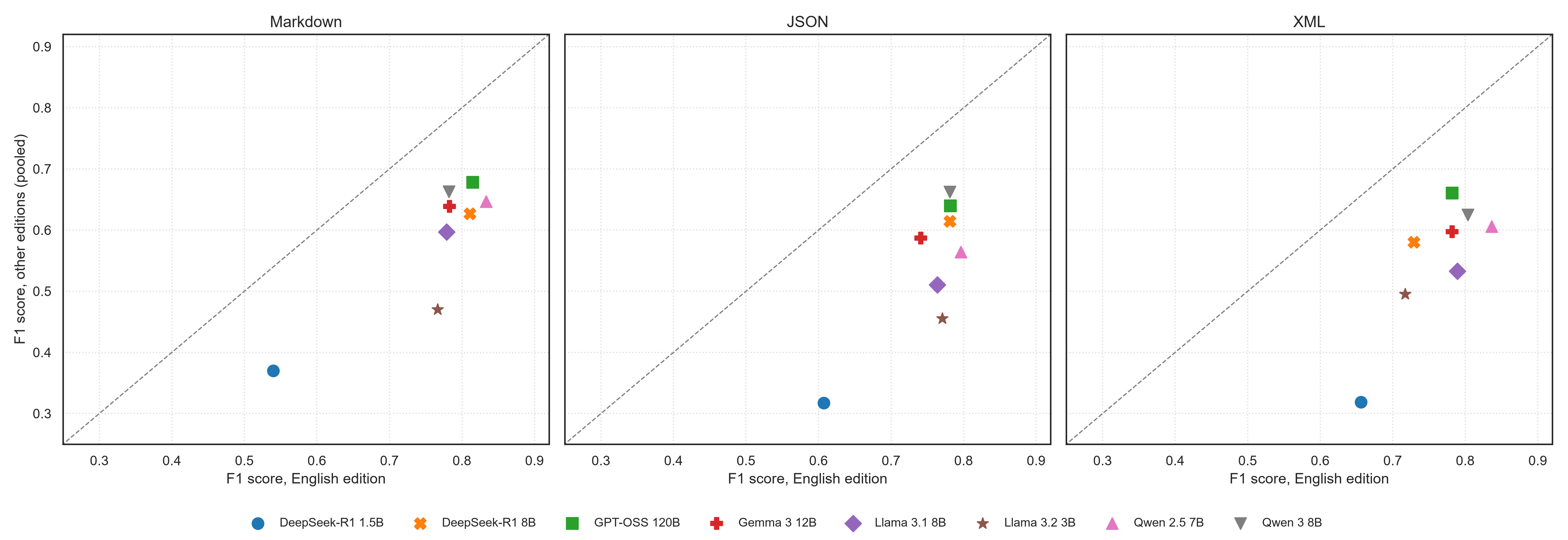}
    \caption{F1 score for the English edition and for the five other editions pooled, by model and format; points below the diagonal indicate lower scores for the other editions.}
    \label{fig:cross_f1}
\end{figure*}

The models demonstrated varying capabilities in cross-lingual retrieval. Although English performance was highest across all tested models, we observed a drop in performance when querying manuals in other languages. With Markdown, F1 dropped by 0.12 to 0.30 from the English edition to the other editions pooled; the drop was largest for Llama 3.2 3B (from 0.77 to 0.47), and for the other editions GPT-OSS 120B (0.68) and Qwen 3 8B (0.66) reached the highest scores. Figure \ref{fig:cross_f1} confirms that smaller models performed worse when the query language differs from the document language.

Of all answers, 884 (5.7\%) were written in the language of the manual edition instead of English, most often by Llama 3.1 8B (248) and least often by GPT-OSS 120B (31); these answers were counted as incorrect (Appendix~\ref{sec:appendix_results}).

The retrieval ablation with four models (Appendix~\ref{sec:appendix_ablation}) showed that, without retrieved context, three of the four models declined every question. For the English edition, BM25, semantic and hybrid retrieval did not differ significantly. For the other editions, BM25 placed the gold page (the page that has the correct answer) among the three retrieved chunks for only 9 to 24\% of the questions, compared with 80 to 93\% for semantic retrieval, and its F1 was 0.35 to 0.47 lower than that of semantic and hybrid retrieval. With the gold page given as context (oracle), F1 for the other editions was 0.06 to 0.11 higher than with hybrid retrieval (significant for 2/4 models), and the gap to the English edition narrowed but remained (0.11 to 0.23).

\section{Discussion}
In this section, we interpret the experimental results to define the architectural requirements for offline AI assistants in agricultural machinery. We analyze the trade-offs between network bandwidth and computational constraints. Finally, we propose a verification framework to integrate these stochastic models into safety-critical environments while respecting the constraints of the ISO 11783 standard.

\subsection{Protocol and Formatting}
The results confirm that transferring operator manuals over the CAN bus is feasible, provided that the system uses an efficient format. We focus on the standard CAN bus to ensure compatibility with the existing vehicle fleet. The bandwidth constraints of ISO 11783 systems make verbose object models (like hierarchical, indented JSON) less attractive. Our results demonstrate that Markdown is highly effective, offering high transfer speeds (47 s vs 83 s for JSON) by minimizing structural metadata while preserving semantic hierarchy. Answer quality did not depend on the format for most models (Section~\ref{sec:llm_performance}), so the choice of Markdown rests on its size, transfer time and readability. JSON had the lowest F1 for most models, significantly for two. The Markdown manuals (317 to 372 kB) are of the same order of magnitude as the Virtual Terminal object pools of commercial implements, as observed by the authors in practice, and their transfer times (47 to 55 s at CTS 32) are comparable to typical object pool uploads.

While Markdown minimizes transmission latency, it shifts complexity to the receiving parser. Unlike XML/JSON, which have mature, MISRA-compliant parsers, Markdown parsing on safety-critical ECUs requires rigorous validation to ensure the reverse text mapping does not drift or hallucinate the document structure. We accept this software complexity trade-off to satisfy the bandwidth constraints of the CAN bus. Furthermore, the presentation layer assumes a Next-Gen Virtual Terminal or a dedicated tablet application, as current ISO 11783-6 object pools possess limited font support and cannot easily render the Unicode characters required for multilingual Markdown display. The additional space required in the Working Set ECU is reasonable for text only, but the number of graphics increases the memory requirements. 

Faster ISOBUS leveraging Ethernet has been proposed. A short-term solution proposal ISO FastLane \citep{oksanen2026isofastlanefasteriso} suggests that with a dual-stack approach, it is possible to accelerate transfer speeds significantly. While that work focused on faster object pool uploads and faster Task Controller rates, it could also accelerate transfer times for operator manuals. 

For standardization, the exchange tested here requires only a new parameter group: the manual is requested with the standard Request message and transferred with ETP, both of which are part of every ISO 11783 stack, and the version check can follow the labels already used to identify Virtual Terminal object pools and Task Controller device descriptions. Because the test relies on a standard stack—including addressing, network management, request handling, and the extended transport protocol—implementing this application for any existing ISO 11783-compliant system is a simple addition to standard features. A standardized, machine-readable manual on the implement would also allow companies other than the implement manufacturer to build products on it, such as assistants, diagnostic tools or training applications.

\subsection{Initialization and Resource Management}
The transfer times were measured under the conditions of the first coupling of an implement (Section~\ref{sec:semsync_methods}), when the system is stationary, as for object pool uploads, and no other large transfer runs at the same time. To keep these conditions, we propose that the Semantic Synchronization protocol be restricted to a safe-state initialization phase. The system should initiate the transfer only when the machine is in a Key-On/Engine-Off state or when the parking brake is engaged. This protocol constraint ensures that the full bandwidth is available for the manual transfer without compromising the safety-critical real-time control loops of the machinery.

While the protocol solves the bandwidth issue, the subsequent vector embedding process represents a significant computational cost for the ECU. We acknowledge that real-time embedding on an embedded CPU is unfeasible during operation. With the first-coupling protocol and the version check (Section~\ref{sec:semsync_methods}), the manual is also embedded only once, and the tractor caches the vector store in non-volatile memory, which avoids both transfer and embedding for daily operations. A CRC32 checksum of the manual is a lightweight way to implement this version check; it detects changes of the manual but does not authenticate it. 

We addressed the trade-off between compression and compute. Although applying compression algorithms (e.g., GZIP) would theoretically equalize the file sizes of JSON and Markdown, we argue against their use in this specific embedded context. Decompressing a large text corpus may require significant CPU cycles. In a Key-On startup scenario, the ECU's CPU is already under heavy load initializing real-time control loops. A raw text format like Markdown avoids this decompression penalty, prioritizing immediate availability over theoretical storage efficiency.

Our analysis reveals a unique opportunity to optimize the system initialization by exploiting the decoupling of computational and network resources. The initialization process consists of two distinct phases: Data Transfer (Network-bound: CAN Bus load) and Vector Embedding (Compute-bound: CPU/NPU load). Because the vector embedding process on embedded hardware is computationally intensive and time-consuming, it creates a window of significant duration where the CAN bus is effectively idle regarding manual data. We therefore propose an opportunistic background transfer of images. Immediately after receiving the text (Markdown), the ECU begins the embedding process. Simultaneously, the system initiates a low-priority background transfer of the image library. Because the embedding process does not consume network bandwidth, these two tasks run in parallel. By the time the embedding phase is complete and the system is ready to answer user queries, the system will have transferred a large portion (or all) of the image library. An on-demand transfer of single images (5.8 s for the sample image, Section~\ref{sec:semsync_results}) then serves only as a fallback mechanism for any images that were not yet captured during the background phase. This approach is intended to avoid the startup delay for text and to reduce the loading latency for images.

We clarify that this architecture implements logical multimodality rather than visual processing. The 8B-parameter models lack the computational capacity for runtime Vision Transformer (ViT) inference. Therefore, the system retrieves the Markdown image reference (e.g., \texttt{![Diagram](id)}) based on the semantic similarity of the surrounding text or explicit image descriptions. This approach avoids the high computational cost of running vision transformers on the ECU while allowing the verification interface to display visual safety data.
\subsection{Context Preservation}
Because standard RAG chunking strategies (e.g., 200-token sliding windows) often sever the text from its parent section headers, retrieving a chunk alone fails to provide the operator with the necessary document context. The proposed self-describing architecture, however, solves this efficiently through reverse text mapping. To generate the UI breadcrumbs shown in our conceptual mockup (Figure \ref{fig:search_navigator}), the system would perform a recursive backward search from the matched chunk to identify the preceding H1, H2, and H3 headers within the Markdown source. Markdown allows for header-based anchoring with zero database overhead. Because the tractor terminal holds the complete source Markdown file in memory, it can perform an exact string search to locate the retrieved chunk within the full document (Appendix~\ref{sec:appendix_ablation}). This operation is computationally negligible (microseconds) for manual-sized documents.

This capability highlights a secondary advantage of Markdown over JSON: Markdown preserves the linear, human-readable text stream. To display the context in a JSON-based system, the terminal would need to reconstruct the document flow from the hierarchical object tree. With Markdown, the terminal simply renders the file, performs a string match for the retrieved answer, scrolls to that location, and highlights it. This feature allows the operator to instantly verify the AI's advice against the surrounding safety warnings and diagrams.

\subsection{Hardware Outlook}
We acknowledge that deployment on legacy automotive ECUs is unfeasible on the tractor side, which in our concept is the consumer of the operator manual data, while the tested concept should work directly for legacy ECUs on the implement side. However, the emergence of next-generation automotive compute platforms could render this architecture viable. The assumption that agricultural vehicles have available embedded nodes is realistic. ECU suppliers have recently introduced high-performance computers (HPCs) in industrially manufactured vehicle platforms. Platforms such as the NVIDIA DRIVE Thor and the Qualcomm Snapdragon Cockpit Elite both feature the native 4-bit hardware acceleration and high-bandwidth unified memory architectures required to run the models we tested locally.

\subsection{Safety and Verification}
\begin{figure*}[pos=tp]
    \centering
    \includegraphics[width=\textwidth]{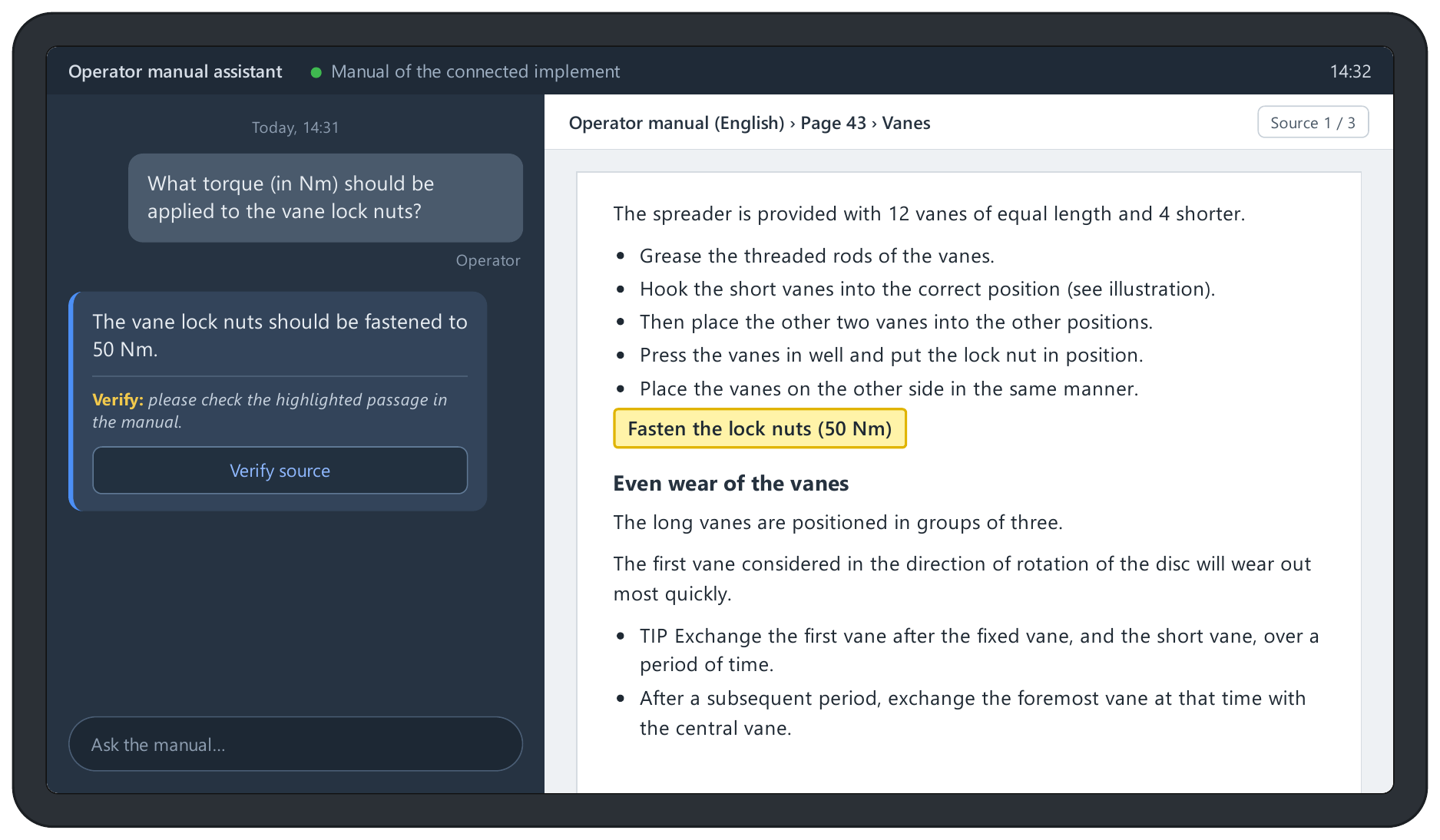}
    \caption{Conceptual mockup of the proposed verification interface. The interface displays the generated answer (left) alongside the original source text (right) for verification by the operator.}
    \label{fig:search_navigator}
\end{figure*}

To enable the reverse text mapping shown in Figure \ref{fig:search_navigator}, the receiving ECU must implement a robust Markdown syntax tree parser to reliably identify parent headers from retrieved text chunks, whereas JSON provides this hierarchy explicitly. Although 8B-parameter models demonstrate high retrieval accuracy, the risk of hallucination—particularly regarding numerical safety data like torque settings—cannot be eliminated. Furthermore, deploying 4-bit quantized weights on edge processors introduces known risks of numerical degradation. Quantization clipping can silently corrupt specific float values generated by the model, potentially altering a 50 Nm torque specification to 5 Nm \citep{gong2025surveylowbitlargelanguage}. This hardware-level limitation elevates the proposed verification interface from an interface convenience to a critical safety requirement. The system relies entirely on the exact-string reverse mapping of the source Markdown to present the original numerical data alongside the generated text. We acknowledge that providing a generated answer introduces the risk of automation bias, where operators may overlook the source text. The verification interface is intended to reduce this risk, not to remove it. 

Therefore, we propose that an agricultural RAG system support verification of its answers: the system guides the operator to the source text instead of acting as the final authority. As proposed in Figure \ref{fig:search_navigator}, the system should not present the LLM's output as an authoritative end-state. Instead, the \textit{Verify Source} trigger opens a split-screen view, aligning the generative response with the primary safety documentation. This change shifts the role of the LLM from an oracle to a navigator, so that the operator can visually verify the AI's advice against the displayed source text. Rather than acting as an authoritative oracle, the proposed interface implements a human-in-the-loop verification layer, a design pattern intended to maintain the operator's role as the primary decision-maker, consistent with safety principles for information systems.

We acknowledge a residual risk regarding visual salience. If the retriever highlights an incorrect section (e.g., tire pressure instead of torque), the operator might succumb to automation bias and trust the highlighted section without reading the text. The architecture prioritizes the verbatim rendering of the source Markdown, so that the operator can base the final decision on the source text.

We argue that the ISOBUS network is a medium that keeps the operator manual physically attached to the machine throughout its lifecycle. Unlike cloud solutions, which depend on connectivity, or USB sticks, which are easily lost, the ECU-stored manual travels with the steel. This local storage ties the documentation to the specific serial number configuration and prevents the drift often observed in cloud systems where manuals document newer hardware revisions than the physical machine possesses. Consequently, the operator has the documentation that matches the machine, regardless of ownership changes.

\subsection{Model Capabilities}
If a model cannot perform the task on high-performance hardware, it will fail on an embedded ECU. Although this study uses PC-grade inference, current ISO 11783-6 Virtual Terminals typically lack dedicated NPU acceleration. Therefore, we propose that this architecture requires a dedicated AI compute node on the CAN bus, separate from the standard Virtual Terminal. The 8B-parameter models would require 4-bit quantization (e.g., GGUF/EXL2) and approximately 5 to 8 GB of VRAM. 

The performance differences between LLMs highlight that the 120B-parameter model offered no advantage over 7B and 8B models such as Qwen 2.5 and Qwen 3 in this task (Section~\ref{sec:llm_performance}), which makes smaller models candidates for running these assistants on edge devices within the tractor cabin. Users must, however, exercise caution. As noted in our previous work \citep{oldpaper},
 hallucination remains a risk, particularly for unanswerable questions. The number of hallucinations of each model is given in Table~\ref{tab:error_categories}.

Cross-lingual capabilities are promising but imperfect with the tested LLMs. For safety-critical applications, native-language manuals are preferable to relying solely on the translation abilities of the LLM. Although high-performing models can bridge the language gap effectively, the drop in performance we observed indicates that accuracy degrades compared to English questions on the English edition. The retrieval ablation shows that semantic retrieval was essential for the other editions, because keyword retrieval failed with English questions on the translated text (Section~\ref{sec:crosslingual_results}). Even with the gold page as context, a gap of 0.11 to 0.23 to the English edition remained, so the remaining drop is not caused by retrieval alone. Part of it came from answers written in the language of the manual instead of English, which also occurred with the gold page (22 to 53 of 540 answers per model).

\subsection{Limitations}

The study used the operator manual of one fertilizer spreader. The six language editions are translations of the same manual; they test cross-lingual behaviour but are not independent validation sets. The 108 questions were written by the authors, their balance of answerable and unanswerable questions does not reflect the questions of operators in practice, and the same questions were asked for every edition, so the observations are correlated. All questions were posed in English, so native-language use was not tested.

The answers were judged automatically. The judge agreed with four human raters for 93.5\% of a stratified sample (Appendix~\ref{sec:appendix_judge_validation}), but its verdicts on individual answers can still be wrong. Each configuration was run once at temperature 0; repeated runs and other software or hardware versions can produce different outputs (up to 0.04 F1 in this study, Appendix~\ref{sec:appendix_ablation}). The retrieval ablation covered four models, the Markdown editions and one set of chunking parameters (200 tokens, 100 tokens overlap, three chunks) and all models used the same embedding model.

Inference ran on PC-grade hardware. Latency, memory use, energy consumption and real-time behaviour on an ECU suitable for the tractor cabin were not tested, nor was the effect of 4-bit quantization on retrieval accuracy and numerical values. The transfers were measured once per configuration on a laboratory bus at the low background load of the first coupling of an implement. Images were not retrieved by their content.

The results refer to the tested serialization (indented JSON, XML without item types, Markdown); minified, compressed or binary representations were not evaluated. The PDF manuals were converted with a wrapper using Docling and used as converted. The proposed verification interface was not evaluated with operators, the derivation of the heading path from a located chunk was not tested, the fidelity of the PDF conversion was not quantified, and integrity and security aspects such as manipulated manuals or prompt injection were outside the scope of this study. 

\section{Conclusions}
In this study, it was demonstrated that a complete operator manual can be transferred over ISO 11783 with the standard request message and the extended transport protocol in 31 to 89 s, which is within the time of a typical object pool upload at the first coupling of an implement. Among the three formats, Markdown produced the smallest files, the shortest transfer times and the fewest tokens, while answer quality was comparable across the formats for most models. Locally run open-weight LLMs answered operator questions from the transferred manual: for the English edition, the models from 3B parameters upward reached F1 scores of 0.77 to 0.83, without an advantage of the 120B model, and for the other editions the scores were 0.12 to 0.30 lower. Semantic retrieval was essential across languages, because keyword retrieval failed for the translated editions, and the hybrid combination gave no advantage over semantic retrieval alone; even with the gold page, part of the drop remained, partly because the models answered in the language of the manual.

Two design elements were also proposed but not evaluated in this study: a first-coupling protocol with a version check, so that the manual is transferred and embedded only once per implement revision, and an on-demand or background transfer of images, for which a single compressed image was transferred in 5.8 s. The next step is a benchmark of the assistants on ECU hardware suitable for the tractor cabin.

\section*{Glossary}
\begin{description}[font=\bfseries]
    \item[CAN] Controller Area Network
    \item[ECU] Electronic Control Unit
    \item[ETP] Extended Transport Protocol
    \item[GNSS] Global Navigation Satellite System
    \item[ISOBUS] ISO 11783 standard
    \item[LLM] Large Language Model
    \item[PGN] Parameter Group Number
    \item[RAG] Retrieval-Augmented Generation
\end{description}

\section*{Acknowledgements}
The authors thank the Kverneland Group for providing the open-access manuals used in this study. We thank especially Peter van der Vlugt for his support to get insight for these publicly available operator manual documents. 

\section*{Data Availability Statement}
The ManualSync code and the question-answer dataset are available at \url{https://github.com/TUMAMX/agriquery/tree/main/ManualSync}.
\section*{Declaration of generative AI and AI-assisted technologies in the manuscript preparation process}
During the preparation of this work the authors used Gemini 3 Pro Preview and Gemini 3 Flash Preview to paraphrase the manuscript text, and Claude (Anthropic) to assist in improving the text and the code. After using these tools, the authors reviewed and edited the content as needed and take full responsibility for the content of the published article.
\bibliographystyle{cas-model2-names}
\bibliography{references}

\appendix
\section{Appendix: QA Test Dataset}
\label{sec:appendix_qa_dataset}

\subsection{Answerable Question Samples}

\begin{itemize}
    \item Q: What torque (in Nm) should be applied to the vane lock nuts? A: 50 Nm
    \item Q: What is the required grease level in mm below the filler opening for spreading disc gearboxes after the machine has stood still? A: 35 mm
    \item Q: Where is the main switch button to turn the control box on or off located? A: The main switch button is located on the upper left in the red extension.
    \item Q: How to enable fine application for dosing low application rate? A: Move the fine application handle to the fine dosing position on both sides.
    \item Q: Where is the RS 232 connector located? A: At the back of the control box.
    \item Q: How often should the agitator axle seal be replaced? A: Every season and after every 100 operational hours.
    \item Q: From which machine point is the spreading height measured to the ground or the crop? A: Measured from the bottom of the vanes.
    \item Q: What materials are required to perform the tray test? A: A measuring tape or ruler, a spirit level, 7 troughs, 7 graduated tubes, a funnel, a notebook, pen, calculator, this manual, and the software's instruction manual.
    \item Q: How many parts does the distribution meter have? A: Seven. 
    \item Q: Should the parking brake of the tractor be engaged before connecting the machine? A: Yes. 
    \item Q: How long should the main switch button be pressed to turn the control box on or off? A: At least 1 second. 
    \item Q: What is the overlap percentage for the full field spreading pattern? A: 100\% overlap. 
    \item Q: What determines the machine's working width? A: Spreading disc RPM
    \item Q: When shortening a coupling shaft, how far must profiled tubes at least overlap in mm? A: 150 mm. 
\end{itemize}

\subsection{Unanswerable Question Samples}
\begin{itemize}
    \item Q: How much extra diesel does the tractor consume to use the Exacta-TLX GEOSPREAD? A: Not found in context
    \item Q: Is one-sided boundary spreading suitable for small gardens? A: Not found in context
    \item Q: Can IsoMatch Tellus be connected to an external mouse? A: Not found in context
    \item Q: Can the linkage pin for the tractor be made out of aluminium? A: Not found in context
    \item Q: What is the maximum height the fertilizer flies when spreading without GEOCONTROL headland? A: Not found in context
    \item Q: What happens if the machine grease nipples are never lubricated? A: Not found in context
    \item Q: What kind of protective safety gloves are needed for cleaning fertiliser remnants from the Exacta-TLX GEOSPREAD before welding? A: Not found in context
    \item Q: What specific 'grease' type is recommended for 'profiled tubes' of the  coupling shaft? A: Not found in context
    \item Q: What is the minimum 'baud rate' for 'RS 232 connection'? A: Not found in context
    \item Q: What is the recommended tire pressure 'range' for transport mode? A: Not found in context
    \item Q: What is the drain rate in liters per minute of 'drain kit' for hopper emptying? A: Not found in context
    \item Q: How many hours of continuous operation can the IsoMatch Tellus operate before flattening a typical tractor battery if left switched on with the engine off? A: Not found in context
    \item Q: What specific paint should be used to paint any damaged paintwork at the end of the season the machine in preparation for winter storage? A: Not found in context
    \item Q: Can the IsoMatch universal ISOBUS terminal be used to check the weather? A: Not found in context
\end{itemize}

\section{Appendix: Prompts}
\label{sec:appendix_prompts}

\subsection{Question prompts}
\begin{appendixpromptstyle}
\texttt{<purpose>} Extract a precise, concise answer to the question from the given context. Adhere strictly to the instructions. Base your answer on the context. \texttt{</purpose>}\par
\texttt{<instructions>}\par
    \hspace*{1em}\texttt{<instruction>} Read the entire context carefully \texttt{</instruction>}\par
    \hspace*{1em}\texttt{<instruction>} Focus ONLY on the specific information related to the question \texttt{</instruction>}\par
    \hspace*{1em}\texttt{<instruction>} Provide an extremely precise answer \texttt{</instruction>}\par
    \hspace*{1em}\texttt{<instruction>} Match the expected answer format exactly \texttt{</instruction>}\par
    \hspace*{1em}\texttt{<instruction>} If unsure, respond with "Unknown" or "Not found in context" \texttt{</instruction>}\par
	\hspace*{1em}\texttt{<instruction>} Answer in English \texttt{</instruction>}\par
\texttt{</instructions>}\par
\vspace{1ex}

\texttt{<context>}\par
    \hspace*{1em}\highlightblue{{context}}\par
\texttt{</context>}\par
\texttt{<question>}\par
    \hspace*{1em}\highlightblue{{question}}\par
\texttt{</question>}\par
\texttt{<answer>}\par
    \hspace*{1em}[Carefully extract the EXACT information that directly answers the question, keeping it as brief and precise as possible]\par
\texttt{</answer>}
\end{appendixpromptstyle}

\subsection{Evaluation Prompt}
\begin{appendixpromptstyle}
\texttt{<purpose>} ANSWER COMPARISON TASK. Do ANSWER\_ONE and ANSWER\_TWO convey the same information regarding the QUESTION? Adhere strictly to the INSTRUCTIONS. Base your ANSWER on the CONTEXT. \texttt{</purpose>}\par
\texttt{<INSTRUCTIONS>}\par
    \hspace*{1em}\texttt{<instruction>} - Respond 'yes' if ANSWER\_ONE and the ANSWER\_TWO convey the SAME TECHNICAL MEANING \texttt{</instruction>}\par
    \hspace*{1em}\texttt{<instruction>} - Consider 'yes' if differences are INSIGNIFICANT to the core technical content \texttt{</instruction>}\par
    \hspace*{1em}\texttt{<instruction>} - Respond 'no' ONLY if there are MEANINGFUL differences that alter the technical understanding \texttt{</instruction>}\par
    \hspace*{1em}\texttt{<instruction>} - Assess the SUBSTANCE of the information, not surface-level variations \texttt{</instruction>}\par
    \hspace*{1em}\texttt{<instruction>} - Answer ONLY with yes or no \texttt{</instruction>}\par
    \hspace*{1em}\texttt{<instruction>} - Don't provide additional information \texttt{</instruction>}\par
\texttt{</INSTRUCTIONS>}\par
\vspace{1ex}

\texttt{<CONTEXT>}\par
    \hspace*{1em}\texttt{<QUESTION>}\par \highlightblue{{question}}\par \texttt{</QUESTION>}\par
    \hspace*{1em}\texttt{<ANSWER\_ONE>}\par \highlightblue{{model\_answer}}\par \texttt{</ANSWER\_ONE>}\par
    \hspace*{1em}\texttt{<ANSWER\_TWO>}\par \highlightblue{{expected\_answer}}\par \texttt{</ANSWER\_TWO>}\par
\texttt{</CONTEXT>}\par
\vspace{1ex}

\texttt{<ANSWER>}\par
    \hspace*{1em}(yes/no)\par
\texttt{</ANSWER>}
\end{appendixpromptstyle}

\section{Appendix: Detailed Test Results}
\label{sec:appendix_results}
\label{sec:appendix_detailed_results}

Table~\ref{tab:f1_detailed} reports the F1 score of every model for each format separately for the English edition and for the five other editions pooled with question-level bootstrap confidence intervals and the scoring rule of Section~\ref{sec:scoring}. 

\begin{table*}[pos=tp]
    \centering
    \caption{F1 score [95\% confidence interval] by model and format.}
    \label{tab:f1_detailed}
    \footnotesize
    \begin{tabular}{llcc}
        \toprule
        Model & Format & English edition & Other editions (pooled) \\
        \midrule
        DeepSeek-R1 1.5B & Markdown & 0.54 [0.45, 0.62] & 0.37 [0.29, 0.45] \\
                         & JSON     & 0.61 [0.51, 0.70] & 0.32 [0.24, 0.39] \\
                         & XML      & 0.66 [0.58, 0.73] & 0.32 [0.24, 0.39] \\
        Llama 3.2 3B     & Markdown & 0.77 [0.68, 0.85] & 0.47 [0.36, 0.57] \\
                         & JSON     & 0.77 [0.69, 0.85] & 0.46 [0.35, 0.56] \\
                         & XML      & 0.72 [0.62, 0.80] & 0.50 [0.39, 0.59] \\
        Qwen 2.5 7B      & Markdown & 0.83 [0.75, 0.90] & 0.65 [0.56, 0.73] \\
                         & JSON     & 0.80 [0.71, 0.87] & 0.56 [0.47, 0.65] \\
                         & XML      & 0.84 [0.77, 0.90] & 0.61 [0.51, 0.70] \\
        Llama 3.1 8B     & Markdown & 0.78 [0.70, 0.85] & 0.60 [0.51, 0.68] \\
                         & JSON     & 0.76 [0.68, 0.84] & 0.51 [0.42, 0.60] \\
                         & XML      & 0.79 [0.71, 0.87] & 0.53 [0.44, 0.62] \\
        DeepSeek-R1 8B   & Markdown & 0.81 [0.74, 0.88] & 0.63 [0.55, 0.70] \\
                         & JSON     & 0.78 [0.70, 0.86] & 0.61 [0.54, 0.69] \\
                         & XML      & 0.73 [0.64, 0.81] & 0.58 [0.50, 0.66] \\
        Qwen 3 8B        & Markdown & 0.78 [0.70, 0.86] & 0.66 [0.58, 0.74] \\
                         & JSON     & 0.78 [0.70, 0.86] & 0.66 [0.58, 0.74] \\
                         & XML      & 0.80 [0.72, 0.88] & 0.62 [0.54, 0.70] \\
        Gemma 3 12B      & Markdown & 0.78 [0.71, 0.85] & 0.64 [0.56, 0.71] \\
                         & JSON     & 0.74 [0.65, 0.82] & 0.59 [0.50, 0.66] \\
                         & XML      & 0.78 [0.70, 0.85] & 0.60 [0.51, 0.68] \\
        GPT-OSS 120B     & Markdown & 0.81 [0.74, 0.89] & 0.68 [0.60, 0.75] \\
                         & JSON     & 0.78 [0.70, 0.85] & 0.64 [0.56, 0.72] \\
                         & XML      & 0.78 [0.70, 0.85] & 0.66 [0.59, 0.73] \\
        \bottomrule
    \end{tabular}
\end{table*}

Table~\ref{tab:error_categories} breaks down the answers of each model into outcome categories, pooled over the six language editions and the three formats. For answerable questions, an incorrect answer was classified as a false refusal, if it stated that the information was not available; and otherwise as a wrong answer. Answers written in the language of the operator manual edition, instead of English, are listed separately whatever the verdict of the judge.

\begin{table*}[pos=tp]
    \centering
    \caption{Outcome categories per model (number of answers; 1944 per model, 972 each to answerable and unanswerable questions). Answers in the language of the operator manual are counted for both question types.}
    \label{tab:error_categories}
    \footnotesize
    \begin{tabular}{lcccccc}
        \toprule
         & \multicolumn{3}{c}{Answerable questions} & & \multicolumn{2}{c}{Unanswerable questions} \\
        \cmidrule(lr){2-4} \cmidrule(lr){6-7}
        Model & Correct & \shortstack{Wrong\\answer} & \shortstack{False\\refusal} & \shortstack{Language\\of manual} & \shortstack{Correct\\abstention} & Hallucination \\
        \midrule
        DeepSeek-R1 1.5B & 367 & 420 & 161 & 45  & 410 & 541 \\
        Llama 3.2 3B     & 419 & 104 & 340 & 160 & 784 & 137 \\
        Qwen 2.5 7B      & 545 & 123 & 190 & 147 & 806 & 133 \\
        Llama 3.1 8B     & 521 & 158 & 118 & 248 & 702 & 197 \\
        DeepSeek-R1 8B   & 557 & 232 & 140 & 59  & 756 & 200 \\
        Qwen 3 8B        & 609 & 216 & 104 & 53  & 748 & 214 \\
        Gemma 3 12B      & 590 & 195 & 97  & 141 & 682 & 239 \\
        GPT-OSS 120B     & 649 & 211 & 90  & 31  & 691 & 272 \\
        \bottomrule
    \end{tabular}
\end{table*}

Of all answers, 884 (5.7\%) were written in the language of the manual edition, all of them for the five non-English editions. The LLM judge accepted 355 of them, because their content matched the expected answer. Scoring these answers by the verdict of the judge (instead of counting them as incorrect) would raise the pooled F1 score for the non-English editions by 0.01 to 0.10 (depending on the model). The results for the English edition are unaffected.

\section{Appendix: Transfer Protocol and Measurements}
\label{sec:appendix_sequence_protocol}

\subsection{Data Transfer Performance Data}

Table \ref{tab:etp_results} presents the measurements from the ISO 11783 ETP transfer tests. 
The tests were carried out with two Windows desktop PCs, each connected to the physical bus with a Kvaser Leaf Light v2 USB-CAN interface and running an in-house C implementation of addressing, request handling and ETP on the Kvaser CANlib library. The bus had a 3 m backbone with stubs of at most 0.5 m and a terminating bias circuit at each end, and it also carried an AEF-certified commercial Virtual Terminal server, Task Controller server and file server. Each combination was transferred once, and transfer times were taken from server-side timestamps. 

The results highlight the effect of the Clear to Send (CTS) parameter; increasing CTS from 16 to 128 packets per burst for the Dutch manual reduced transfer time from 66.6 s to 30.8 s. The indented JSON files (Section~\ref{sec:formatting_analysis}) required the longest transfer times. The compressed sample image (\texttt{image\_test}) was transferred in 5.8 s.

\begin{table*}[pos=tp]
\centering
\footnotesize
\caption{Detailed ETP data transfer results comparing languages, formats, and protocol settings (CTS). "DT" refers to Data Transfer packets; "CM" refers to Connection Management packets.}
\label{tab:etp_results}
\begin{tabular}{lrrrrrr}
\toprule
\textbf{Filename} & \textbf{Size (B)} & \textbf{DT Pkts} & \textbf{CM Pkts} & \textbf{Load} & \textbf{Time} & \textbf{Param}  \\
\midrule
dutch\_manual.md & 319540 & 45649 & 1 + 2855 & 41.1\% & 66.6 s & CTS 16 \\
dutch\_manual.md & 319540 & 45649 & 1 + 1428 & 55.6\% & 47.1 s & CTS 32 \\
dutch\_manual.md & 319540 & 45649 & 1 + 715 & 70.9\% & 35.4 s & CTS 64 \\
dutch\_manual.md & 319540 & 45649 & 1 + 358 & 79.1\% & 30.8 s & CTS 128 \\
\midrule
dutch\_manual.xml & 334470 & 47782 & 1 + 1495 & 55.5\% & 48.9 s & CTS 32 \\
dutch\_manual.json & 555565 & 79367 & 1 + 2482 & 54.8\% & 84.2 s & CTS 32 \\
\midrule
english\_manual.md & 317454 & 45351 & 1 + 1419 & 55.3\% & 47.0 s & CTS 32 \\
english\_manual.xml & 333699 & 47672 & 1 + 1491 & 54.9\% & 49.3 s & CTS 32 \\
english\_manual.json & 556063 & 79438 & 1 + 2484 & 55.6\% & 82.7 s & CTS 32 \\
\midrule
french\_manual.md & 371708 & 53102 & 1 + 1661 & 55.6\% & 54.9 s & CTS 32 \\
french\_manual.xml & 387402 & 55344 & 1 + 1731 & 55.4\% & 57.0 s & CTS 32 \\
french\_manual.json & 607023 & 86718 & 1 + 2711 & 56.8\% & 89.1 s & CTS 32 \\
\midrule
german\_manual.md & 347323 & 49618 & 1 + 1552 & 55.9\% & 50.7 s & CTS 32 \\
german\_manual.xml & 363334 & 51905 & 1 + 1624 & 55.4\% & 53.1 s & CTS 32 \\
german\_manual.json & 585542 & 83649 & 1 + 2616 & 56.4\% & 86.4 s & CTS 32 \\
\midrule
italian\_manual.md & 360976 & 51568 & 1 + 1613 & 55.4\% & 52.9 s & CTS 32 \\
italian\_manual.xml & 377485 & 53927 & 1 + 1687 & 55.0\% & 56.0 s & CTS 32 \\
italian\_manual.json & 597026 & 85290 & 1 + 2667 & 57.4\% & 86.3 s & CTS 32 \\
\midrule
spanish\_manual.md & 358426 & 51204 & 1 + 1602 & 55.6\% & 52.2 s & CTS 32 \\
spanish\_manual.xml & 373605 & 53373 & 1 + 1669 & 55.5\% & 55.0 s & CTS 32 \\
spanish\_manual.json & 591478 & 84497 & 1 + 2642 & 56.5\% & 87.0 s & CTS 32 \\
\midrule
image\_test & 39855 & 5694 & 1 + 179 & 41.4\% & 5.8 s & CTS 32 \\
\bottomrule
\end{tabular}
\end{table*}

\subsection{Format Efficiency}
\begin{figure}[pos=htbp]
    \centering
    \includegraphics[width=\columnwidth]{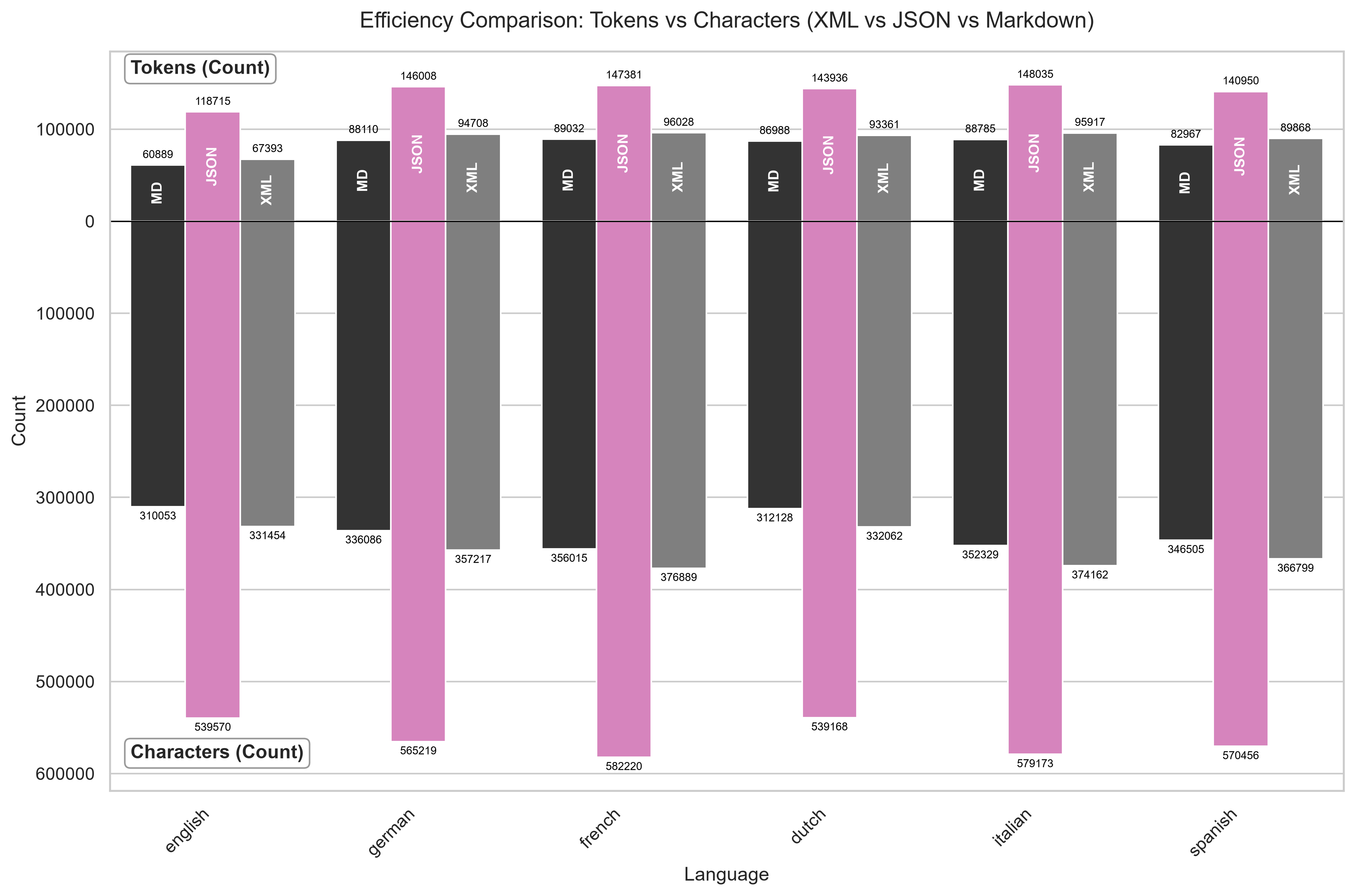}
    \caption{Token and character counts of the converted manuals by format and language edition (tokens: Qwen3-Embedding-8B tokenizer).}
    \label{fig:token_counts_chars}
\end{figure}

\section{Appendix: Retrieval Ablation}
\label{sec:appendix_ablation}

The retrieval configuration of RAG was evaluated in two parts: a) a retrieval-only evaluation without a language model and b) an end-to-end ablation with five context conditions. Our earlier study with the same question set compared keyword, semantic and hybrid retrieval with another embedding model \citep{oldpaper}; the hybrid configuration, which performed best there, was retained for the main experiments of this study.

\subsection{Retrieval-only evaluation}
For the 54 answerable questions, the chunks of the main experiments were ranked by BM25; by semantic search and by hybrid fusion (RRF with $k = 60$ and ten candidates per retriever). A retrieved chunk counted as a hit, if it covered the gold page of the question; each chunk was mapped to its pages through the page markers of the converted operator manual. For this mapping, each chunk was located by exact string search in its converted manual; all chunks of the 18 collections were found verbatim. Table~\ref{tab:retrieval_metrics} reports Recall@3, the share of questions, for which the gold page was among the three chunks (@3) passed to the model.

\begin{table*}[pos=tp]
    \centering
    \caption{Recall@3 of the gold page for the 54 answerable questions (English questions on each language edition).}
    \label{tab:retrieval_metrics}
    \footnotesize
    \begin{tabular}{l ccc ccc ccc}
        \toprule
         & \multicolumn{3}{c}{Markdown} & \multicolumn{3}{c}{JSON} & \multicolumn{3}{c}{XML} \\
        \cmidrule(lr){2-4} \cmidrule(lr){5-7} \cmidrule(lr){8-10}
        Edition & BM25 & Semantic & Hybrid & BM25 & Semantic & Hybrid & BM25 & Semantic & Hybrid \\
        \midrule
        English & 0.89 & 0.93 & 0.89 & 0.91 & 0.91 & 0.94 & 0.80 & 0.91 & 0.89 \\
        German  & 0.11 & 0.87 & 0.78 & 0.13 & 0.91 & 0.78 & 0.11 & 0.89 & 0.78 \\
        French  & 0.24 & 0.83 & 0.80 & 0.19 & 0.89 & 0.83 & 0.19 & 0.87 & 0.74 \\
        Dutch   & 0.20 & 0.83 & 0.78 & 0.19 & 0.83 & 0.80 & 0.17 & 0.93 & 0.85 \\
        Italian & 0.09 & 0.83 & 0.80 & 0.09 & 0.87 & 0.78 & 0.09 & 0.83 & 0.78 \\
        Spanish & 0.15 & 0.80 & 0.76 & 0.17 & 0.87 & 0.74 & 0.13 & 0.89 & 0.76 \\
        \bottomrule
    \end{tabular}
\end{table*}

For the English edition of the manual, the three methods retrieved the gold page for 80 to 94\% of the questions. For the five other editions: BM25 retrieved it for only 9 to 24\% of the questions, whereas the semantic search retrieved it for 80 to 93\%. In these editions, hybrid fusion retrieved the gold page less often than the semantic search alone (74 to 85\%), because chunks ranked only by BM25 displaced semantic hits in the fused ranking. The mean reciprocal rank of the first chunk from the gold page within the top ten (MRR@10) showed the same pattern.

\subsection{End-to-end Ablation}
Five context conditions were compared with four models (Llama 3.2 3B, Qwen 2.5 7B, DeepSeek-R1 8B and Gemma 3 12B) on the Markdown editions of all six languages, with the question prompt and the judge of the main experiments: 1) no context (empty context), 2) BM25 only, 3) semantic only, 4) hybrid (our configuration of the main experiments) and 5) oracle (the text of the gold page; for unanswerable questions, the thematically related page given in the dataset). 

F1 scores were computed with the scoring rule of Section~\ref{sec:scoring}, with question-level bootstrap confidence intervals; paired contrasts were corrected with the Holm method (Table~\ref{tab:ablation}).

\begin{table*}[pos=tp]
    \centering
    \caption{F1 score [95\% confidence interval] by context condition (Markdown editions; other languages pooled).}
    \label{tab:ablation}
    \footnotesize
    \setlength{\tabcolsep}{4pt}
    \begin{tabular}{l ccccc}
        \toprule
        Model & No context & BM25 & Semantic & Hybrid & Oracle \\
        \midrule
        \multicolumn{6}{l}{\textit{English edition}} \\
        Llama 3.2 3B   & 0.11 [0.03, 0.21] & 0.74 [0.64, 0.83] & 0.75 [0.67, 0.84] & 0.76 [0.66, 0.84] & 0.80 [0.72, 0.88] \\
        Qwen 2.5 7B    & 0.00 [0.00, 0.00] & 0.81 [0.72, 0.88] & 0.81 [0.73, 0.89] & 0.81 [0.73, 0.88] & 0.87 [0.81, 0.93] \\
        DeepSeek-R1 8B & 0.00 [0.00, 0.00] & 0.73 [0.63, 0.82] & 0.76 [0.67, 0.85] & 0.77 [0.67, 0.85] & 0.84 [0.76, 0.91] \\
        Gemma 3 12B    & 0.00 [0.00, 0.00] & 0.78 [0.69, 0.85] & 0.82 [0.75, 0.89] & 0.78 [0.71, 0.86] & 0.84 [0.78, 0.90] \\
        \midrule
        \multicolumn{6}{l}{\textit{Other five editions (pooled)}} \\
        Llama 3.2 3B   & 0.16 [0.06, 0.27] & 0.15 [0.07, 0.24] & 0.55 [0.44, 0.64] & 0.50 [0.39, 0.60] & 0.57 [0.46, 0.66] \\
        Qwen 2.5 7B    & 0.00 [0.00, 0.00] & 0.19 [0.10, 0.29] & 0.66 [0.57, 0.74] & 0.63 [0.54, 0.72] & 0.74 [0.66, 0.82] \\
        DeepSeek-R1 8B & 0.00 [0.00, 0.00] & 0.20 [0.11, 0.29] & 0.62 [0.55, 0.70] & 0.64 [0.57, 0.71] & 0.70 [0.63, 0.76] \\
        Gemma 3 12B    & 0.00 [0.00, 0.00] & 0.22 [0.12, 0.31] & 0.67 [0.59, 0.74] & 0.64 [0.56, 0.72] & 0.73 [0.66, 0.80] \\
        \bottomrule
    \end{tabular}
\end{table*}

Without context, three models declined every question and Llama 3.2 3B answered 11\% of the answerable questions correctly; hybrid retrieval therefore raised F1 by 0.65 to 0.81 for the English edition and by 0.34 to 0.64 for the other editions (all $p < 0.001$ after Holm correction). 

For the English edition, the retrieval methods did not differ significantly. For the other editions, semantic and hybrid retrieval exceeded BM25 by 0.35 to 0.47 (all $p < 0.001$), whereas the hybrid and the semantic retrieval did not differ significantly. 

The gold page raised F1 above hybrid retrieval by 0.06 to 0.11 for the other editions, significantly for Gemma 3 12B and Qwen 2.5 7B. With the gold page, the gap between the English edition and the other editions was reduced from 0.13--0.26 (hybrid) to 0.11--0.23 (oracle) so that part of the gap remained when the retrieval was not a factor.

\section{Appendix: Validation of the LLM Judge}
\label{sec:appendix_judge_validation}

The automatic evaluation (judging) relied on GPT-OSS 20B as judge (Section~\ref{sec:scoring}). To understand the accuracy of this LLM-as-judge, its verdicts were validated against human judgement, as reported here. The items for which the LLM returned no valid verdict were judged again.

For 147 of the 15552 items of the main experiments (0.9\%), the judge returned no valid verdict: 136 outputs were empty, because the reasoning of the judge reached the output limit of 1024 tokens before an answer was produced, and 11 outputs read ``(no)'' instead of ``no''. These items were judged again with identical settings, up to five times; 123 items received a valid verdict (105 at the first and 18 at the second repetition). The remaining 24 items were judged with the reasoning effort of the judge set to low, which resolved all of them. The same procedure was applied to the ablation experiments (Appendix~\ref{sec:appendix_ablation}). No item was therefore excluded from the analysis.

Four independent non-author raters assessed a stratified sample of 210 items. The raters had basic understanding of the machinery type in question, but were not experts on this particular machine or its operator manual. 

The sample comprised six items from each of the 32 combinations of the eight models, the two question types (answerable, unanswerable) and the two judge verdicts (yes, no) together with 18 of the 24 items judged with low reasoning effort (the other six were answers in the language of the manual or mixed answers, which were not sampled). Answers in the language of the operator manual edition (884 items) and answers mixing English with that language (226 items) were not sampled. This was done because the scoring rule counts the former as incorrect regardless of the verdict and the latter contain passages, that the raters could not be expected to read. For each item, the raters saw a) the question, b) the expected known answer and c) the model answer. The model, the language edition, the format or the verdict of the judge were not shown to the raters. They chose one of three options: match, no match or cannot decide. A verdict of the judge was compared with the majority of at least 3/4 raters.

The raters agreed substantially (Fleiss' $\kappa = 0.83$, 95\% confidence interval 0.77 to 0.88); a three-of-four majority was reached for 201 of 210 items. The judge agreed with this majority for 188 of the 201 items (93.5\%; Cohen's $\kappa = 0.87$, 95\% confidence interval 0.80 to 0.93; Table~\ref{tab:judge_confusion}). Weighted to the composition of all evaluated answers, the accuracy of the judge was 0.96 (95\% confidence interval 0.93--0.98).

\begin{table}[pos=htbp]
    \centering
    \caption{Verdicts of the judge compared with the majority of four human raters (201 items with a majority).}
    \label{tab:judge_confusion}
    \footnotesize\setlength{\tabcolsep}{3pt}
    \begin{tabular}{lcc}
        \toprule
         & Human majority: match & Human majority: no match \\
        \midrule
        Judge: yes & 102 & 3 \\
        Judge: no  & 10 & 86 \\
        \bottomrule
    \end{tabular}
\end{table}

Most disagreements were rejections of answers that the raters accepted (10 of 13). In the stratified part of the sample (186 items with a majority), for answerable questions, the judge was wrong in 15.6\% of its negative verdicts (7 of 45) and in 4.3\% of its positive verdicts (2 of 46); for unanswerable questions, in 2.1\% of its negative verdicts (1 of 48) and in none of its positive verdicts (0 of 47). 

Correcting these errors would raise the pooled F1 score of the English and language-neutral answers from 0.64 to about 0.67, so the reported F1 scores are conservative. For the items judged with low reasoning effort, the judge agreed with the majority for 12 of 15 items. Disagreements among the raters concerned partially correct or loosely worded answers (such as incomplete lists, missing specific details and paraphrased locations).

\end{document}